\documentclass[runningheads]{llncs}
\usepackage{graphicx}
\usepackage{times}
\usepackage{latexsym}
\usepackage{multirow}
\usepackage{booktabs}
\usepackage{textcomp}
\usepackage{threeparttable}
\usepackage{amsmath}
\usepackage{float}
\usepackage{amssymb}
\usepackage{arydshln}
\usepackage{todonotes}
\usepackage{color, colortbl}
\usepackage{makecell}

\begin{document}
\title{An Information Extraction Study: Take In Mind the Tokenization!\vspace{-7mm}}
%
%
\author{Christos Theodoropoulos\inst{1,2}\and
Marie-Francine Moens\inst{1,2}\vspace{-3mm}}
\authorrunning{C. Theodoropoulos, M.F. Moens}
%
\institute{KU Leuven, Oude Markt 13, 3000 Leuven, Belgium
\email{christos.theodoropoulos@kuleuven.be}, \email{sien.moens@kuleuven.be} \and Language Intelligence and Information Retrieval Lab, Celestijnenlaan 200A, \\3001 Leuven, Belgium} 

\maketitle              
\begin{abstract}
\vspace{-8mm}
Current research on the advantages and trade-offs of using characters, instead of tokenized text, as input for deep learning models, has evolved substantially. New token-free models remove the traditional tokenization step; however, their efficiency remains unclear. Moreover, the effect of tokenization is relatively unexplored in sequence tagging tasks. To this end, we investigate the impact of tokenization when extracting information from documents and present a comparative study and analysis of subword-based and character-based models. Specifically, we study Information Extraction (IE) from biomedical texts. The main outcome is twofold: tokenization patterns can introduce inductive bias that results in state-of-the-art performance, and the character-based models produce promising results; thus, transitioning to token-free IE models is feasible.
\vspace{-4mm}
\vspace{-5mm}
\end{abstract}
\section{Introduction}
\vspace{-3mm}
Currently, neural network models are replacing traditional Natural Language Processing (NLP) pipelines, as their ability to learn abstract and meaningful representations improves the performance. Hence, the complex and error-prone handcrafted feature engineering has been substantially reduced. However, the word-level or subword-level tokenization step remains, being carried over from the traditional era of NLP systems. Designing a custom tokenizer based on linguistic characteristics is time-consuming, expensive, and language specific and requires feature engineering and linguistic-related expertise. To alleviate these issues, data-driven approaches, such as WordPiece \cite{wu2016google}, Byte Pair Encoding \cite{sennrich2016neural}, and SentencePiece \cite{kudo2018sentencepiece}, tokenize the text by splitting the strings based on frequent words and subwords (word pieces) given a corpus. Nonetheless, these algorithms struggle to handle special linguistic morphologies \cite{clark2021canine} and their impact in sequence tagging tasks, such as Named Entity Recognition (NER), is relatively unexplored. The open research discussion on the tokenization step motivates the first research question of the study:

\begin{itemize}
    \vspace{-2.5mm}
    \item How does the tokenization step affect the performance in the IE task? (RQ1) 
    \vspace{-2.5mm}
\end{itemize}

In this paper, we conduct a tokenization analysis and inductive bias study to explore the existence of potential patterns, related to the tokenization step, by solving the IE task in the biomedical domain using subword-based models. We refer to the general definition of inductive bias in AI models, as the set of assumptions that the learner uses to predict outputs of given inputs that it has not encountered \cite{mitchell1980need}. We base our study on Partition Filter Network (PFN) \cite{yan-etal-2021-partition}, which solves the NER \cite{nadeau2007survey,florian-etal-2010-improving}  and Relation Extraction (RE) \cite{sun-etal-2011-semi,plank-moschitti-2013-embedding} tasks jointly by modeling the interaction between the tasks and learning independent and shared representations. We choose this model because it leverages the Language Model (LM) representations by design. Hence, we can experiment with different LMs and explore their effectiveness. The main outcome of the analysis is that the tokenization patterns introduce inductive bias in the IE task. Additionally, the similarity analysis of the learned entity representations probes the existence of inductive bias. Following this key finding, we explore the capabilities of a tokenization bias-free model and answer the second research question:

\begin{itemize}
    \vspace{-3mm}
    \item Can a transition to character-level models be carried out without significant performance degradation? (RQ2)
    \vspace{-3mm}
\end{itemize}

Recently, new character-based models \cite{el-boukkouri-etal-2020-characterbert,clark-etal-2022-canine,xue-etal-2022-byt5,tay2021charformer} that directly process sequences of characters have been released, and transitioning to this kind of model by replacing subword-based models without losing performance has become a focus of research. Hence, we conduct a comparative study for the IE task, including subword-based and character-based models. Additionally, we present a hyphenation analysis to detect possible linguistic characteristics, by exploring patterns of subwords with a length of 4 characters, and probe the hypothesis that character-based models are more capable of capturing special text morphology.

In summary, the key contributions are as follows:

\begin{itemize}
    \vspace{-3.5mm}
    \item We present an extensive analysis to investigate the effect of tokenization in the IE task for the biomedical domain and raise awareness.
    \item We identify the existence of inductive bias when specific tokenization patterns are detected, which leads to new state-of-the-art (SOTA) performance in the ADE dataset \cite{gurulingappa2012development}.
    \item We present a comparative study, including subword-based and character-based models, and draw insights supported by the hyphenation analysis.
    \vspace{-6mm}
\end{itemize}

\section{Tokenization Analysis - Datasets}
\vspace{-4mm}
In this section, we conduct a tokenization analysis for the dataset used in the study. We choose the biomedical (ADE) dataset to explore the effect of tokenization in a special domain. The ADE dataset contains entities of Drugs and Adverse Effects (AE) and has labels for the relations between them. The tokenizer of cased BERT \cite{devlin-etal-2019-bert} and bioclinical BERT (b-BERT) \cite{alsentzer-etal-2019-publicly} is based on the WordPiece algorithm, while ALBERT \cite{lan2019albert} adopts the SentencePiece algorithm.

In Tab. 2, we present the effect of tokenization on the average sentence length, in terms of word pieces (subwords), for each dataset. The sentence length increases by approximately 12 tokens, up to 58\%, after the tokenization in the biomedical domain. To further explore the number of word pieces per entity type, we isolate the unique entities\footnote{We note that the set of unique entities for the case and uncased text processing is different, which is why the initial average entity length might be different.}. Then, we find the unique words that are part of each entity type and tokenize the unique entities and words using the different tokenizers to notice the difference in the length and the addition of the word pieces. In Tab. 1 the last column represents the average tokenized word length per entity type, and the \textit{Out} type describes the words that are not part of an entity of interest. In the ADE dataset, the length of the drug and AE entities increases substantially, and the drug entities are split into more word pieces. Particularly, a word that is part of a drug entity is split into approximately 4 word pieces, on average, when using the tokenizer of cased BERT and b-BERT. The tokenizer of ALBERT tends to split the entities of interest into fewer pieces.
\vspace{-3mm}

\begin{table}[!t]
    \parbox{.5\linewidth}{
    \centering
          \caption{Average Entity Length - ADE dataset}
          \centering
          \resizebox{\linewidth}{!}{
          \begin{threeparttable}
          \begin{tabular}{lcccc}
            \toprule
            Tokenizer & Type & Entity & Tokenized Entity\tnote{1} & Word\\
            \midrule
            cased BERT & \multirow{3}{*}{Drug} & 1.37 & 4.78 (+248.9\%) & 3.92\\
            b-BERT &    & 1.37 & 4.79 (+249.6\%) & 3.93\\
            ALBERT &    & 1.42 & 4.37 (+207.7\%) & 3.38\\
            \midrule
            cased BERT & \multirow{3}{*}{AE} & 2.66 & 6 (+125.6\%) & 2.81\\
            b-BERT &    & 2.66 & 5.9 (+121.8\%) & 2.77\\
            ALBERT &    & 2.72 & 5.29 (+94.5\%) & 2.38\\
            \midrule
            cased BERT & \multirow{3}{*}{Out} & -- & -- & 2.11\\
            b-BERT &    & -- & -- & 2.06\\
            ALBERT &    & -- & -- & 2.09\\
          \bottomrule
          \end{tabular}
          \begin{tablenotes}
              \item [1] (+ x \%): percentage increase
          \end{tablenotes}
          \end{threeparttable}}
    }
    \hfill
    \parbox{.5\linewidth}{
    \centering
          \caption{Average Sentence Length}
          \centering
          \resizebox{\linewidth}{!}{
          \begin{threeparttable}
          \begin{tabular}{lccc}
            \toprule
            Tokenizer & Dataset & Sentence & Tokenized Sentence\tnote{1}\\
            \midrule
            cased BERT & \multirow{3}{*}{ADE} & \multirow{3}{*}{21.23} & 33.56 (+58.1\%)\\
            b-BERT &      &      & 33.1 (+55.9\%)\\
            ALBERT &      &      & 33.25 (+56.6\%)\\
           \bottomrule
          \end{tabular}
          \begin{tablenotes}
              \item [1] (+ x \%): percentage increase
          \end{tablenotes}
          \end{threeparttable}}
    }
    \vspace{-6mm}
\end{table}

\section{Inductive Bias}
\vspace{-3mm}
In this section, we answer the first research question (RQ1). Following the observations of the tokenization analysis, we conduct a study to investigate whether tokenization patterns introduce inductive bias in the IE task. 
\vspace{-3.5mm}
\subsection{Experimental Setup}
\vspace{-2.5mm}
The overall model architecture of this paper is presented in Fig. 1. The sentence is processed by an LM, followed by an aggregation step that constructs the word-level embeddings by calculating the summed and averaged representations. When the aggregation step is not used, the model operates in the subword level as the PFN module directly processes the output of the LM. The PFN module models a two-way interaction between the NER and RE tasks, as it leverages the representations of the LM and segments the neurons into two task partitions (independent representations) and one shared partition (inter-task interaction). PFN consists of a partition filter encoder, a NER unit, and a RE unit \cite{yan-etal-2021-partition}. The partition filter encoder is a recurrent feature encoder that stores information in intermediate memories. In each step, the neurons of the encoder are divided into three partitions: the relation, entity, and shared partitions. Then the encoder combines these partitions for task-specific feature generation and filters out irrelevant, for each task, information \cite{yan-etal-2021-partition}. The NER-specific and RE-specific features are the input of the NER and RE units respectively. In this section, we focus on the subword-based LMs (cased BERT, b-BERT, ALBERT XLL) and run experiments with and without the aggregation step to explore differences in performance. The models are trained end-to-end.

We train the PFN module\footnote{All the experiments are executed using a GeForce RTX 3090 24GB GPU.} (Fig. 1) using the hyperparameters that are selected in the official paper of the model \cite{yan-etal-2021-partition} to solve the joint IE task. The training epochs are set to 100, the batch size is 20, and the learning rate is 2e-5. We use ADAM \cite{kingma2014adam} as the optimizer and keep the best model based on the performance in the development set in each run. For the ADE dataset, 10-fold cross-validation\footnote{We use the same split as \cite{eberts2020span}.} is applied \cite{li2017neural,bekoulis2018joint,eberts2020span}, and 15\%\footnote{Random split with the same seed for a fair comparison.} of the training set is used as the development set. We use strict evaluation of the IE task \cite{bekoulis2018joint,taille2020let}. An entity is predicted correctly if the boundaries and the type are detected. A relation is correct if the type and the two involved entities are predicted correctly. We conduct a statistical t-test (p-value $\leq$ 0.05) of the evaluation results to draw conclusions in a more robust manner\footnote{The code and trained models are publicly available in the repository of the paper for reproducibility and to facilitate further research. https://github.com/christos42/inductive\textunderscore bias\textunderscore IE}. The results of the statistical t-test are available in the Appendix section. We highlight that the inductive bias study is based on the intra-model comparison, as we focus on the effect of the aggregation step. 

\begin{figure}[!h]
\vspace{-7mm}
  \centering
  \includegraphics[width=0.8\textwidth]{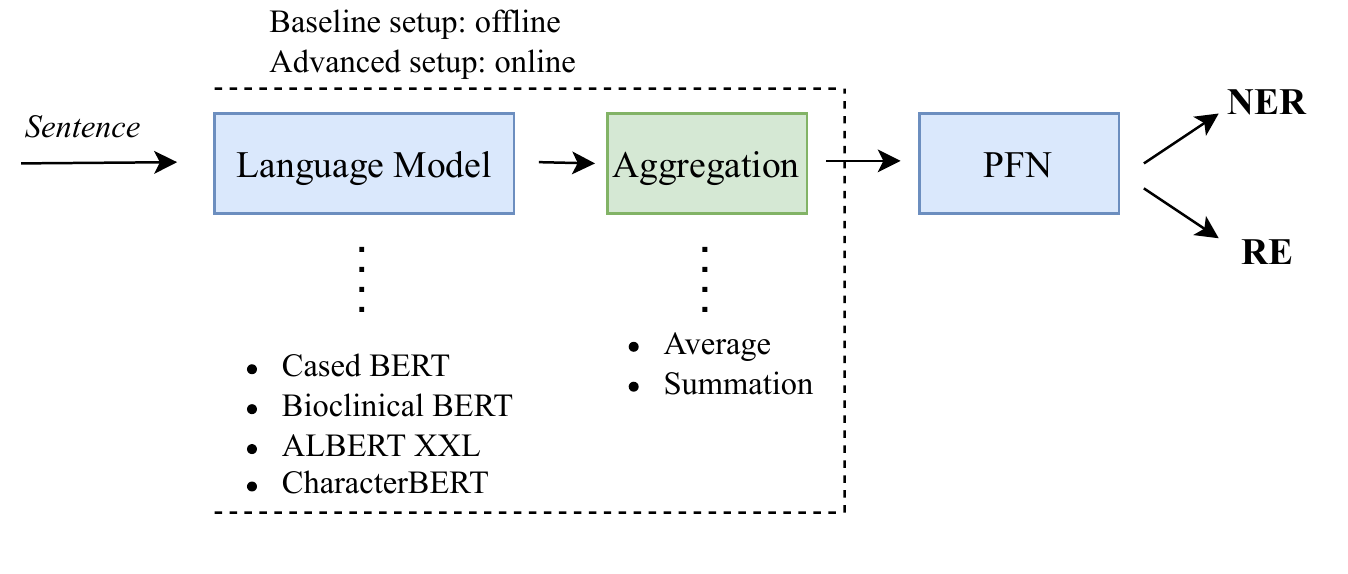}
   \vspace{-8mm}
   \caption{Model Architecture: The input sentence passes through the language model and then the embeddings are aggregated if needed. Finally, the representations are the input of the PFN module and the final predictions for the NER and RE task are extracted.}
   \vspace{-12mm}
\end{figure}

\subsection{Results - Discussion}

\begin{table*}[!h]
    \vspace{-7.5mm}
    \caption{End-to-end training - Results}
    \vspace{-2.5mm}
    \centering
    \begin{tabular}{cccc}
        \toprule
        Language Model & Aggregation & NER & RE\\
        \midrule
        \multirow{3}{*}{cased BERT} & - & 89.2 \textpm{ 1.3}  & 80.2 \textpm{ 2.6} \\
                                    & Average & 89.7 \textpm{ 1.1} & 80.5 \textpm{ 2.3} \\
                                    & Summation & 89.9 \textpm{ 1.1} & 80.5 \textpm{ 2} \\
        \midrule
        \multirow{3}{*}{ALBERT XXL} & - & 90.8 \textpm{ 0.9} & 83.2 \textpm{ 2.1} \\
                                    & Average & \textbf{91.5} \textpm{ 0.8} & \textbf{83.9} \textpm{ 1.6} \\
                                    & Summation & 91.2 \textpm{ 0.8} & 83 \textpm{ 1.3} \\
        \midrule
        \multirow{3}{*}{b-BERT} & - & 89.6 \textpm{ 1} & 81.1 \textpm{ 2.2} \\
                                & Average & 90.1 \textpm{ 1.1} & 81.3 \textpm{ 2.1} \\
                                & Summation & 90.5 \textpm{ 0.9} & 81.9 \textpm{ 2.1} \\
        \midrule
        \midrule
        CharacterBERT & - & 91.2 \textpm{ 1} & 83.2 \textpm{ 1.8} \\
        \bottomrule
    \end{tabular}
    \vspace{-6mm}
\end{table*}

For every model, the aggregation is beneficial as it improves the performance in both NER and RE tasks (Tab. 3). More precisely, the addition of summed aggregation improves the performance by 0.7\%, 0.9\% (NER task), and 0.3\%, 0.8\% (RE task) for the cased BERT-based and b-BERT-based models respectively, compared to the aggregation-free models. For the ALBERT-based model the averaged aggregation boosts the performance by 0.7\% in both tasks. Coupling this finding with the tokenization analysis (Tab. 2), the pattern of word-piece splitting for words of interest (Drugs and AE) acts as inductive bias when aggregation is used. The intra-model comparison reveals the existence of inductive bias since the only difference lies in the addition of the aggregation step. Even if the aggregation layer (simple summation and averaging) is not trainable, the incoming gradient (backpropagation, \cite{rumelhart1986learning}) from the PFN module appears to be more informative for the IE task.
\vspace{-5.5mm}

\subsection{Similarity Analysis}
\vspace{-2.5mm}
An entity can consist of multiple words, and the entity boundaries should be detected correctly by the model. Hence, the initial and the end words of the entity are important. An entity can be split into multiple word pieces. For example, the drug \textit{sodium polystyrene sulfonate} (3 words) is split into \textit{sodium p-oly-sty-rene su-lf-ona-te} (9 word pieces) when the tokenizer of BERT is used. When the aggregation step is not used, the model should detect the initial word (sodium) and the end token (su) of the entity. In the inference step, the correctly detected entity can be reconstructed with detokenization.

To more deeply investigate the inductive bias phenomenon, we conduct a cosine similarity analysis for the different entities. The hypothesis is that the detected inductive bias in the biomedical text can increase the similarity and robustness of the entity representations. We use the trained LM of each run of the inductive bias study, with and without aggregation, and extract the representations of the test set. Then, we separate the words of the entities based on the entity type and the ordering of the words (start/end words). Hence, we have two groups per entity. One contains the start words, and another contains the end words. The \textit{Joint} group consists of both the start and end words. The average similarity of each group is calculated. As we run the experiments using 10-fold cross-validation for the ADE dataset, we average the averaged similarity scores across the different splits. The results discussion is based on the intra-model comparison.

In the ADE dataset (Tab. 4), generally, the averaged entity similarity is increased when aggregation is used. Hence, the detected inductive bias, which is correlated with the tokenization patterns, results in more similar entity representations. In particular, the summed representations of cased BERT and b-BERT are more or almost equally similar compared to the averaged and the aggregation-free representations for both entity types. Especially, for the \textit{Drug} entity, the similarity increment is up to 10.5\%, 3\%, and 14.5\% for the \textit{Start-word}, \textit{End-word}, and \textit{Joint} groups respectively, when aggregation  is used. Accordingly, for the \textit{Adverse-Effect} entity, the similarity increase is up to 11.5\%, 4\%, and 13.5\%. For the ALBERT XXL language model, where the tokenization patterns are less profound (Tab. 2), the similarity slightly increases, when aggregation is used, in most cases. However, the increment is significantly lower than that detected with the representations of cased BERT and b-BERT.

\begin{table}[!b]
    \vspace{-6mm}
    \caption{Similarity Analysis: Average cosine similarity scores per entity group. The total average scores across the different experimental runs are presented.}
    \vspace{-2.5mm}
    \centering
    \resizebox{0.6\linewidth}{!}{
        \begin{tabular}{lccccccc}
        \toprule
        \multirow{2}{*}{Language Model} & \multirow{2}{*}{Aggregation} & \multicolumn{3}{c}{Drugs} & \multicolumn{3}{c}{Adverse Effects}\\
        \cmidrule{3-8}
        &  & Start & End & Joint & Start & End & Joint\\
        \midrule
        \parbox[t]{3mm}{\multirow{3}{*}{cased BERT}} & --\hspace{3mm} & 67.43 & 82.65 & 56.84 & 51.43 & 78.24 & 40.44\\
                                                        & Average & 68.3 & \textbf{85.73} & 61.15 & 56.78 & \textbf{82.5} & 49.75\\
                                                        & Summation & \textbf{78.02} & 84.5 & \textbf{71.11} & \textbf{61.54} & 81.18 & \textbf{50.89}\\
        \midrule
        \parbox[t]{3mm}{\multirow{3}{*}{b-BERT}} & --\hspace{3mm} & 67.76 & 84.51 & 57.06 & 53.86 & 79.92 & 41\\
                                                        & Average & 68.91 & \textbf{86.01} & 63.3 & 60.84 & \textbf{83.07} & 53.85\\
                                                        & Summation & \textbf{78.12} & 85.52 & \textbf{70.87} & \textbf{65.23} & 81.55 & \textbf{54.63}\\
        \midrule
        \parbox[t]{3mm}{\multirow{3}{*}{ALBERT XXL}} & --\hspace{3mm} & 65.41 & \textbf{82.06} & 57.23 & \textbf{55.46} & 77.79 & 44.62\\
                                        & Average & 65.89 & 81.03 & 57.88 & 55.18 & \textbf{78.46} & \textbf{46.11}\\
                                        & Summation & \textbf{67.53} & 77.06 & \textbf{59.93} & 53.44 & 73.42 & 43.68\\
        \bottomrule
    \end{tabular}
    }
\end{table}

\section{Comparative Study}
\vspace{-4.5mm}
The existence of inductive bias that is related to the initial tokenization of the text motivates the second research question (RQ2) of the paper. When tokenization patterns are present and the likelihood of splitting a word of interest (part of an entity) into multiple word pieces is higher, the addition of an aggregation step increases the performance and the robustness of the entity representations. Since the improved performance is correlated with this kind of inductive bias, a comparison with character-based models that do not include a tokenization step is important.
\vspace{-4.5mm}
\subsection{Baseline Setup: Frozen Embeddings}
\vspace{-3mm}
As we want to explore how feasible the transition to tokenizer-free models is, we categorize the LMs into two categories based on the way they handle the input text: subword-based and character-based models. In the comparative study, we use bioclinical BERT for the ADE dataset\footnote{We use the Transformers library \cite{wolf-etal-2020-transformers}.} to represent the subword-based set of models and we select CharacterBERT \cite{el-boukkouri-etal-2020-characterbert} as character-based representative. CharacterBERT processes the initial text at the character level and removes the tokenization step by incorporating the character-CNN module \cite{peters-etal-2018-deep} to learn representations at the word level. Since we intend to evaluate the significance of tokenization, we include CharacterBERT and BERT models in the study, as their main architecture is identical and their difference lies in the tokenization step. 

In the baseline setup, we want to directly evaluate the quality of the "off-the-shelf" representations of different LMs when solving the IE task. As we directly evaluate the pretrained representations, it is important to mention the corpus that was used for pretraining the different LMs. Bioclinical BERT was initialized with BioBERT (pretrained on PubMed abstracts and PMC OA \footnote{PubMed Central Open Access: https://www.ncbi.nlm.nih.gov/pmc/tools/openftlist/}) \cite{lee2020biobert} parameters and pretrained on MIMIC III notes \cite{johnson2016mimic}. The medical version of CharacterBERT was retrained on MIMIC III notes and PMC OA biomedical article abstracts. Hence, the medical version of CharacterBERT and b-BERT were pretrained with almost identical data and a comparison between these groups of LMs is safe.

First, we extract the word representations of each LM of the study offline. We aggregate the subword-level embeddings (b-BERT) and construct the word-level embeddings by calculating the averaged and summed representations. CharacterBERT extracts word-level representations by design. The overall experimental setup (hyperparameter) is the same as the inductive bias study setting. The only difference is that the LM is frozen (Fig. 1).

\begin{table*}[!h]
    \vspace{-7.5mm}
    \caption{Baseline Setup - Results}
    \vspace{-3.5mm}
    \centering
    \begin{tabular}{lccc}
        \toprule
        Language Model & Aggregation & NER & RE\\
        \midrule
        \multirow{2}{*}{b-BERT} & Average & 85.6 \textpm{ 0.7} & 75.7 \textpm{ 1.7}\\
                                & Summation & 85.7 \textpm{ 0.7} & 75.1 \textpm{ 1.7} \\
        \midrule
        CharacterBERT & - & \textbf{87.5} \textpm{ 0.8}  & \textbf{77.9} \textpm{ 1.5} \\
        \bottomrule
    \end{tabular}
    \vspace{-5mm}
\end{table*}

The model that leverages the representations of medical CharacterBERT performs significantly better, as it outperforms the model that uses the b-BERT representations by around 2\% in both RE and NER tasks (Tab. 5). This finding illustrates that medical CharacterBERT is more capable of exploring and learning the special linguistic characteristics of biomedical text, as it produces more meaningful representations than b-BERT. For the subword-based LM, the two aggregation strategies result in similar performance.
\vspace{-4mm}

\subsection{Advanced Setup: End-to-end Training}
\vspace{-2mm}
In the advanced setup, we conduct experiments with end-to-end training to also fine-tune the LMs and make comparisons on the final performance. To this end, we extend the setup of the inductive bias study and incorporate the character-based model of the comparative study. The CharacterBERT-based model achieves very competitive results and outperforms the b-BERT-based models in the advanced setup. The performance improvement is 0.7\% (NER task) and 1.3\% (RE task) (Tab. 3). This is an additional indication that CharacterBERT is more capable of detecting special linguistic characteristics of medical-related text, despite the suggestion that the subword-based LMs can independently learn the essential character compositions \cite{itzhak2021models}. 
\vspace{-4mm}

\subsection{Hyphenation Analysis}
\vspace{-2mm}
The comparative study reveals that the model that leverages CharacterBERT is very competitive in the biomedical text. Following this main observation, we conduct a hyphenation analysis to explore possible special linguistic characteristics in the biomedical domain. The hypothesis is that the CharacterBERT-based model performs very well because there is domain-specific linguistic morphology in the biomedical text. We find the unique words for each entity type and then we extract all subwords with a length of 4 characters and calculate the frequency of each subword per entity type.

Figs. 2 and 3 present the 25 most frequent subwords, excluding those that are in the 50 most frequent subwords of the out-of-entity words\footnote{The unique words that are not a part of any entity type of the dataset.}, for the \textit{Drug} and \textit{Adverse-Effect} entities respectively. A special morphology is noticeable for both entity types. Specifically, the words that are part of the \textit{Drug} and the \textit{Adverse-Effect} entity have 11 (e.g. \textit{amin}, \textit{mine}, \textit{mide}, etc.) and 19 (\textit{itis}, \textit{osis}, \textit{emia}, etc.) subwords accordingly, with frequencies higher than 20. These findings confirm the initial hypothesis of the hyphenation analysis.

\begin{figure}[!h]
\vspace{-6mm}
  \centering
  \includegraphics[width=0.65\linewidth]{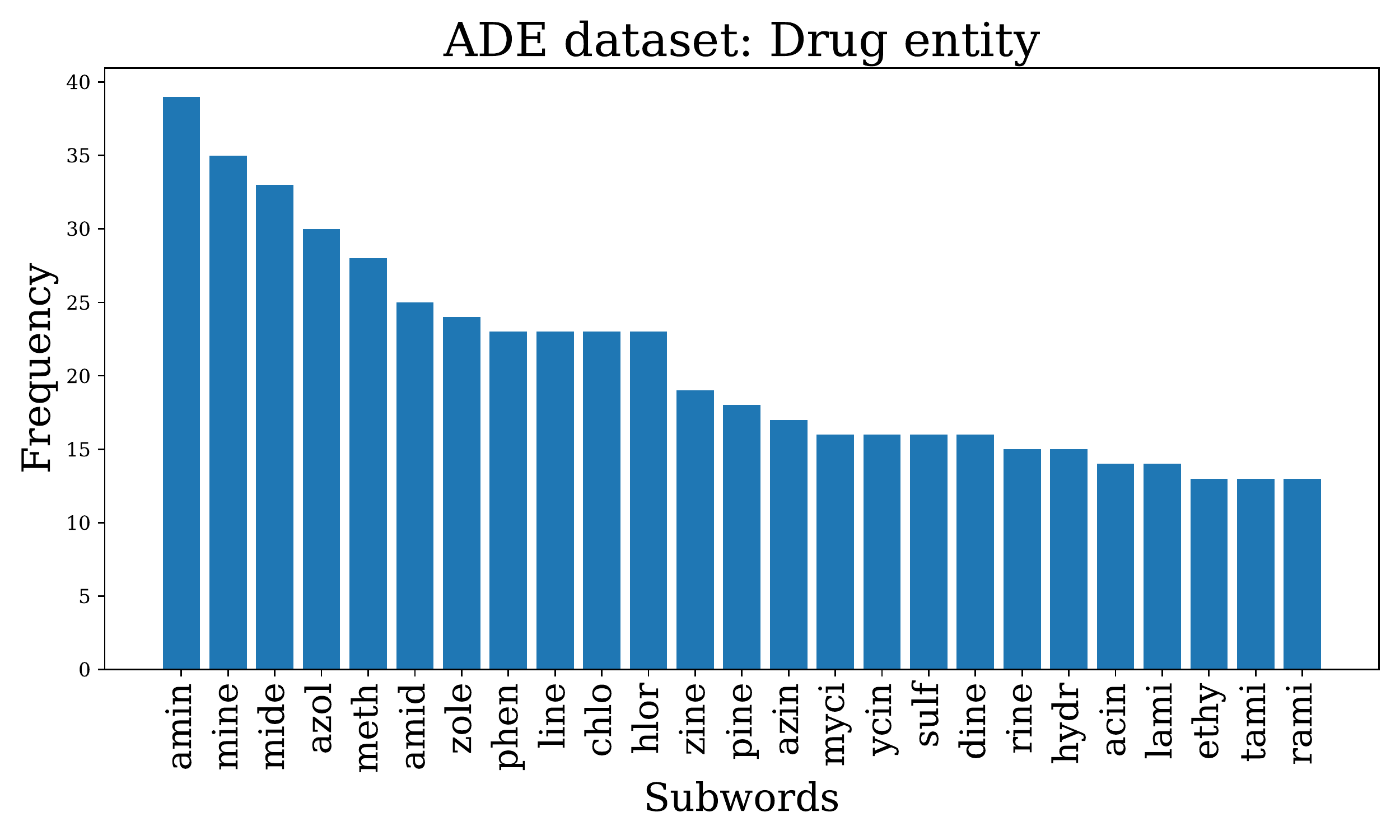}
  \vspace{-6mm}
  \caption{Drug entity: 25 most frequent subwords with a length of 4 characters}
\end{figure}

\begin{figure}[!h]
  \vspace{-4mm}
  \centering
  \includegraphics[width=0.65\linewidth]{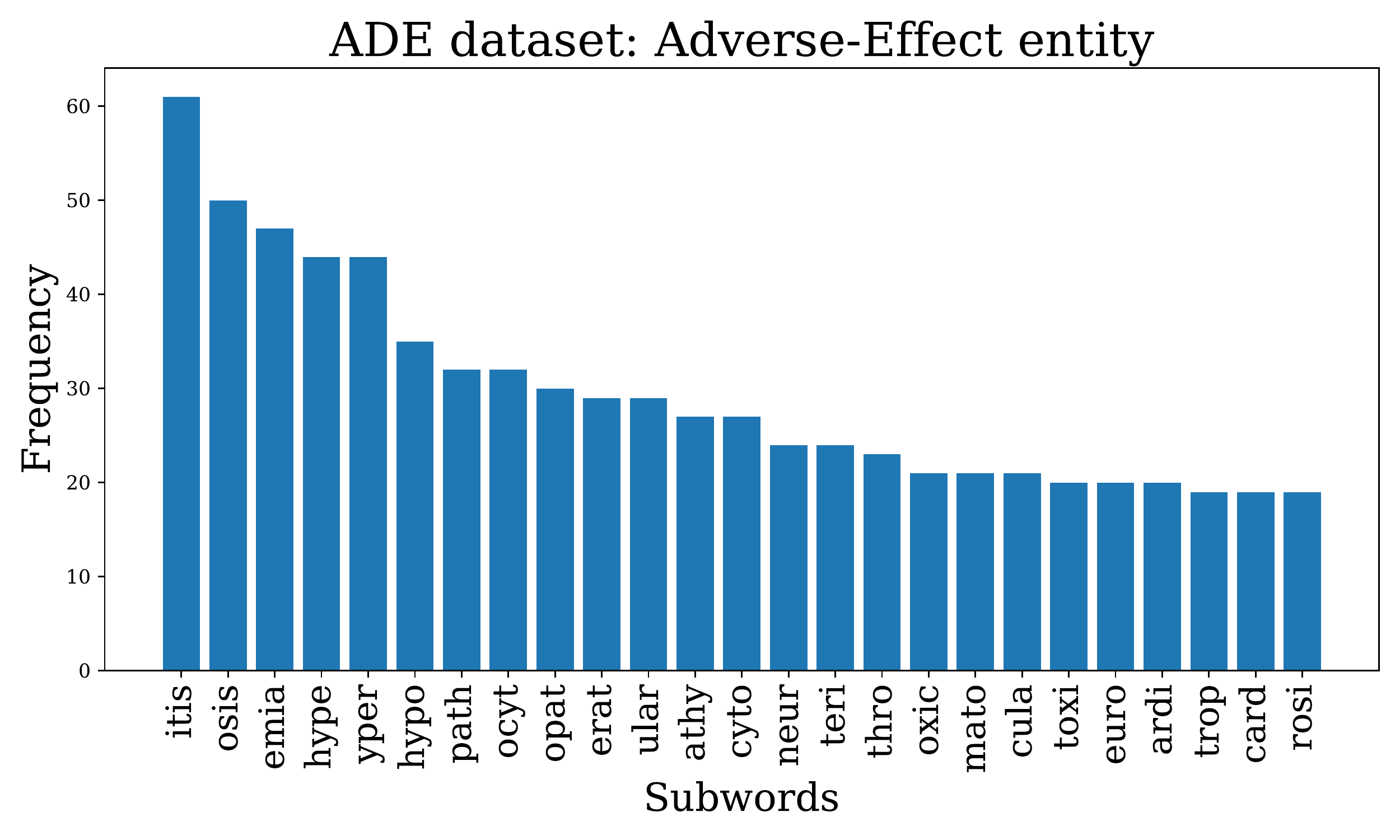}
  \vspace{-6mm}
  \caption{Adverse-Effect entity: 25 most frequent subwords with a length of 4 characters}
  \vspace{-6mm}
\end{figure}

Tab. 6 presents the number of entity subwords with frequencies equal to or higher than a set of thresholds. Noticeable patterns can be detected in the biomedical domain where all of the 25 most frequent subwords for both entities (\textit{Drug} and \textit{Adverse-Effect}) have a higher than 10 frequency.

\begin{table}[!h]
    \vspace{-5mm}
    \caption{Number of entity subwords with frequency higher than a specific threshold, subword length: 4 characters}
    \vspace{-3mm}
    \centering
    \begin{tabular}{ccccc}
        \toprule
        \multirow{2}{*}{Entity type} & \multicolumn{4}{c}{Threshold} \\
        \cmidrule{2-5}
        & $\geq$ 40 & $\geq$ 30 & $\geq$ 20 & $\geq$ 10 \\
        \midrule
        Drug & 0 & 3 & 11 & 25 \\
        Adverse-Effect & 5 & 8 & 19 & 25 \\
        \bottomrule
    \end{tabular}
    \vspace{-9mm}
\end{table}

\section{Comparison With SOTA Models}
\vspace{-2mm}
For comparison, we choose models that are trained on the same dataset without extra external data. In the ADE dataset, we outperform the SOTA models. More precisely, the ALBERT XLL-based model with average aggregation improves the performance by 0.7\% and 0.2\% in the RE and NER task respectively (Tab. 7). The inductive bias that is introduced by the tokenization patterns and is exploited with the aggregation layer boosts the performance. 

\begin{table}[!h]
    \vspace{-7mm}
    \caption{Comparative Results - SOTA}
    \vspace{-3mm}
    \centering
    \begin{tabular}{llcc}
     \toprule
    Dataset & Model & NER & RE\\
    \midrule
    \parbox[t]{2mm}{\multirow{6}{*}{\rotatebox[origin=c]{90}{ADE}}} & Eberts and Ulges (2020) \cite{eberts2020span} & 89.3 & 79.2\\
                                                                    & Theodoropoulos et al. (2021) \cite{theodoropoulos-etal-2021-imposing} & 88.3 & 80.0\\
                                                                    & Wang and Lu (2020) \cite{wang-lu-2020-two} & 89.7 & 80.1\\
                                                                    & Zhao et al. \cite{ijcai2020-558} & 89.4 & 81.1\\
                                                                    & Yan et al. \cite{yan-etal-2021-partition} & 91.3 & 83.2\\
                                                                    \cmidrule{2-4}
                                                                    & ALBERT XXL (Avg. Aggr.), PFN & \textbf{91.5} & \textbf{83.9}\\
    \midrule
    \end{tabular}
    \vspace{-10mm}
\end{table}

\section{Related Work}
\vspace{-3mm}
Ács et al. (2021) \cite{acs-etal-2021-subword} explore the effect of subword pooling on three tasks: morphological probing, POS, and NER tagging. Zhang and Tan (2021) \cite{zhang2021textual} present a comparison of different textual representations for cross-lingual information retrieval. Traditional token \cite{sasaki-etal-2018-cross}, subword \cite{tiedemann2020opus} and character-level representations are compared for the German, French and Japanese languages. The main outcome is that leveraging the traditional token representations results in the best performance, and combining subword representations can be beneficial in some cases. In our study, we compare the pretrained representations of subword-based and character-based LMs in the IE task. In addition, we explore the performance of the different models in an end-to-end training setup.

Itzhak and Levy (2021) \cite{itzhak2021models} discuss models that implicitly learn at the character level even when they are trained on the subword level. More precisely, they explore the capabilities of RoBERTa-base and Large \cite{liu2019roberta}, GPT2-medium \cite{radford2019language}, and AraBERT-large \cite{antoun-etal-2020-arabert} in word spelling. The main results indicate that the embedding layer of the subword-level LMs contains considerable information about the character composition of the tokens. The study does not include character-based LMs (e.g. CharacterBERT) by design. In contrast, we compare subword-based and character-based LMs in the IE task and implicitly explore the capabilities of the models on capturing special linguistic morphology (biomedical text) by presenting a hyphenation analysis.

To the best of our knowledge, there is no related work for the first research question of our paper and the revealing of inductive bias in the IE task when tokenization patterns are present.
\vspace{-4mm}

\section{Conclusion}
\vspace{-2.5mm}
This paper identifies the existence of inductive bias in the IE task that is correlated with tokenization patterns, where the words of interest are more likely to be split into subwords. We highlight the introduction of inductive bias in the biomedical domain, supported by a similarity analysis based on entity representations. Additionally, we conduct a comparative study, including subword-based and character-based models, pointing out that the transition to token-free IE models is achievable. In future work, we intend to explore the effect of tokenization in other sequence tagging problems.
\vspace{-4mm}

\section*{Limitations}
\vspace{-2.5mm}
A limitation of the paper is that the dataset is relatively small. Nevertheless, this is a common problem in the IE field, and in our case, it is beneficial in the sense that we can experiment quickly and run multiple experiments to draw conclusions. If the dataset is large, the computational power needed for the study will increase by a considerable factor. Potentially, additional language models can be incorporated into the comparative study \cite{clark-etal-2022-canine,xue-etal-2022-byt5} but retraining with identical data is needed to alleviate the influence of the different pretraining corpora. Isolating and exploring the effect of tokenization in a comparative inter-model setup is challenging because other factors, such as the different architecture of the language models, can affect the performance. We highlight that, for this reason, we incorporate BERT and CharacterBERT in the comparative study as these models have the same architecture and their only difference lies in the tokenization step. 
\vspace{-8mm}

\section*{Acknowledgments}
This work is supported by the Research Foundation – Flanders (FWO). The authors are affiliated with Leuven.AI - KU Leuven institute for AI, B-3000, Leuven, Belgium. This preprint has not undergone any post-submission improvements or corrections. The Version of Record of this contribution will be published in Lecture Notes in Computer Science (LNCS). The DOI will be mentioned when the publication is finished. 

\bibliographystyle{splncs04}
\bibliography{mybibliography}

%




\end{document}